\documentclass{article}

\usepackage{PRIMEarxiv}

\usepackage[utf8]{inputenc} 
\usepackage[T1]{fontenc}    
\usepackage{hyperref}       
\usepackage{url}            
\usepackage{booktabs}       
\usepackage{amsfonts}       
\usepackage{nicefrac}       
\usepackage{microtype}      
\usepackage{lipsum}
\usepackage{fancyhdr}       
\usepackage{graphicx}       
\graphicspath{{media/}}     
\usepackage{amsmath} 

\usepackage{enumitem}
\usepackage{soul}
\usepackage{multirow}
\usepackage{multicol}
\usepackage{xcolor}
\usepackage{array}

\title{Transfer Learning Approach for Railway Technical Map (RTM) Component Identification}

\author{
  Obadage Rochana Rumalshan \\
  University of Peradeniya  \\
  Peraderniya, Sri Lanka \\
  \texttt{rochanar@sci.pdn.ac.lk} \\
  \And
  Pramuka Weerasinghe  \\
  University of Peradeniya  \\
  Peraderniya, Sri Lanka \\
  \texttt{pramukaw@sci.pdn.ac.lk} \\
    \And
   Mohamed Shaheer  \\
  University of Peradeniya  \\
  Peraderniya, Sri Lanka \\
  \texttt{shaheerm@sci.pdn.ac.lk} \\
  \And
  Prabhath Gunathilake  \\
  University of Peradeniya  \\
  Peraderniya, Sri Lanka \\
  \texttt{prabhathg@sci.pdn.ac.lk} \\
  \And
  Erunika Dayaratna \\
  University of Peradeniya  \\
  Peraderniya, Sri Lanka \\
  \texttt{erunika.dayaratna@sci.pdn.ac.lk} \\
}

\begin{document}
\maketitle

\begin{abstract}
The extreme popularity over the years for railway transportation 
urges the necessity to maintain efficient railway management systems around the globe. Even though, at present, there exist a large collection of Computer Aided Designed Railway Technical Maps (RTMs) but available only in the portable document format (PDF). Using Deep Learning and Optical Character Recognition techniques, this research work proposes a generic system to digitize the relevant map component data from a given input image and create a formatted text file per image. Out of YOLOv3, SSD and Faster-RCNN object detection models used, Faster-RCNN yields the highest mean Average Precision (mAP) and the highest F1 score values 0.68 and 0.76 respectively. Further it is proven from the results obtained that, one can improve the results with OCR when the text containing image is being sent through a sophisticated pre-processing pipeline to remove distortions.
\end{abstract}


\section{Introduction}
Railway Transportation is extremely popular all around the globe and urges the requirement of digitized databases that includes railway track information with all railway track components such as signals, switches and mileposts (Figure 1). A Railway Technical Map (RTM) is a complex diagram (Figure 1) which includes all the information associated with a railway track. At present, most railway companies maintain RTMs designed with computer aided software, yet they are only available in PDF format. These contain partially distorted map components where identifying those components using basic digital image processing techniques is hard due to its complexity. This work focuses on implementing an automated system to generate CSV formatted files for given RTM input images containing all the digitized data that can be used with further decision support tools. The final formatted text will include the component associativity with mileposts, component names and descriptions.

\begin{figure}[h]
\vspace{-0.1cm}
  \centering
    \setlength{\fboxsep}{0pt} 
    \setlength{\fboxrule}{0.5pt} 
  \fbox{\includegraphics[width=1\linewidth]{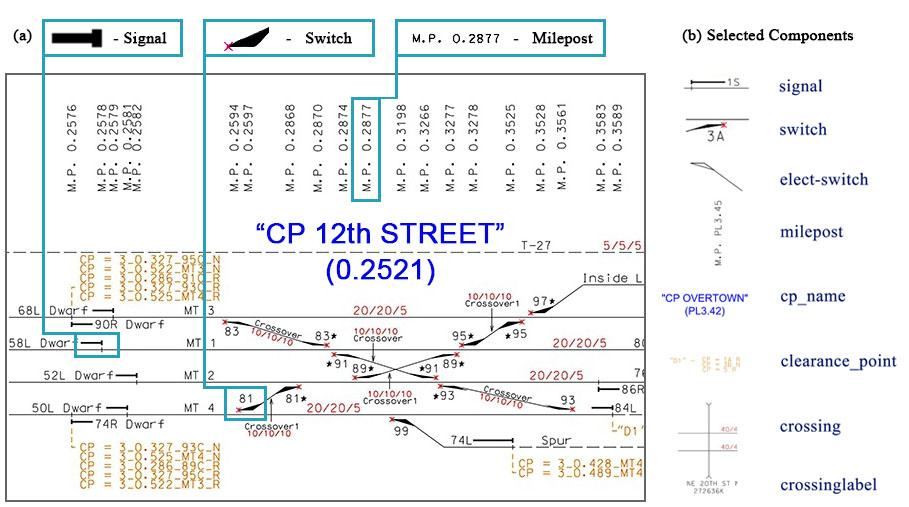}}
  \caption{(a) A Railway Technical Map (b) Selected components for labelling.}
  \label{Railway-Technical-Map-1}
\vspace{-0.3cm}
\end{figure}

\section{Background and Literature}

Jiao and their colleagues descriptively describe the available state of the art Object Detection models \cite{8825470-ref-1} which includes YOLOV3, SSD, SPP-net, Faster-RCNN, R-FCN, Mask-RCNN and Rotation Mask-RCNN. Among the few diagram digitizing solutions proposed in the literature, Moreno-Garcı´a’s team presents a general framework for the digitization of complex Engineering Drawings \cite{Moreno-García2019-ref-2}. Even though there are numerous researches on object detection and optical character recognition, it seems that only available process or a method for identification of components and associated data in RTM using a Deep Learning (DL) approach, is the work carried out by our team \cite{EasyChair:8456-ref-3}. Still the components mapping with corresponding mileposts was not discussed there.

\section{Methodology}

\subsection{Approach}
\begin{figure}[h]
\vspace{-0.1cm}
  \centering
    \setlength{\fboxsep}{0pt} 
    \setlength{\fboxrule}{0.5pt} 
  \fbox{\includegraphics[width=1\linewidth]{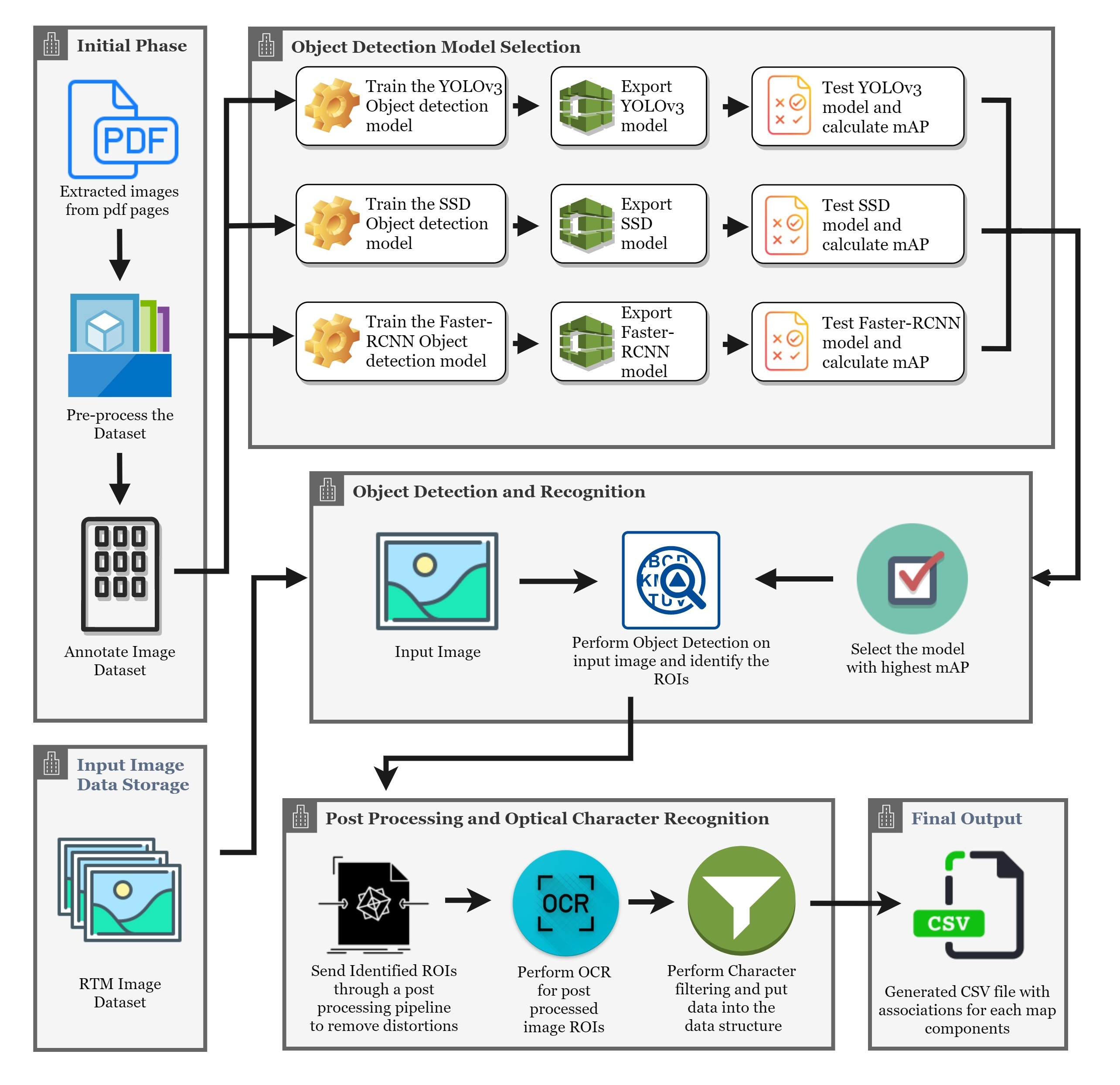}}
  \caption{Complete System Architecture}
  \label{Complete System Architecture}
\vspace{-0.3cm}
\end{figure}

All DL models SSD, YOLOv3 and Faster-RCNN, are tested to distinguish the best fit with highest mean Average Precision (mAP) value, which is then used as the candidate to detect components in the input images. After sending the identified regions of interest through a digital image pre-processing pipeline to remove grid-like distortions, those are sent to the OCR process. After the OCR, the text that is read, is mapped with its respective milepost where the component is located in actual context and stored temporarily using an internal data structure. Once the OCR process is completed for all components of the given image, the stored data will be used to produce the final output CSV file containing digitized data for detected components as shown in the Figure \ref{Complete System Architecture} complete system architecture.

\subsection{Dataset}
As the initial dataset 69 PDF pages (ex: Figure \ref{Railway-Technical-Map-1} (a)) of RTMs which are been originally constructed using a Computer Aided Design (CAD) tool, are received that representing the range of a single mile in each page. These are converted from PDF to JPEG format and pre-processed to reduce the size of images to 4500px width with 2400px height for the ease of object detection model training attempts.


\subsection{Image labelling}
The pre-processed dataset contains the 8 different components namely Signal, Switch, Electric-Switch, Milepost, Control Point Name, Clearance Point, Crossing and Crossing Label (Figure \ref{Railway-Technical-Map-1} (b)). “labelImg” \cite{githubGitHubHumanSignallabelImg-ref-4} is used as the image annotation tool.

\subsection{Model Development}
The initial dataset is split as train and test sets in 2:1 ratio, respectively. For the training environment, “Google Colaboratory” free online GPU service is used. The dataset is trained on an NVIDIA Tesla K80 GPU with 12GB of RAM. Since the scarcity of training data due to confidentiality concerns, Transfer Learning approach is used. 

\subsection{Optical Character Recognition}
The results obtained from object localizations through the object detection model with highest mAP, are used for OCR to extract and identify the text areas and con-tent that are associated with identified component. To remove unwanted areas and other distortions, the concepts, Seeded Region Growing, Masking with bit-wise operations are employed and a set of character lists are defined and the final text produced text produced for each detection is filtered using the above stated character lists.

\subsection{Enhancing OCR}
\begin{figure}[h]
\vspace{-0.1cm}
  \centering
    \setlength{\fboxsep}{0pt} 
    \setlength{\fboxrule}{0.5pt} 
  \fbox{\includegraphics[width=0.8\linewidth]{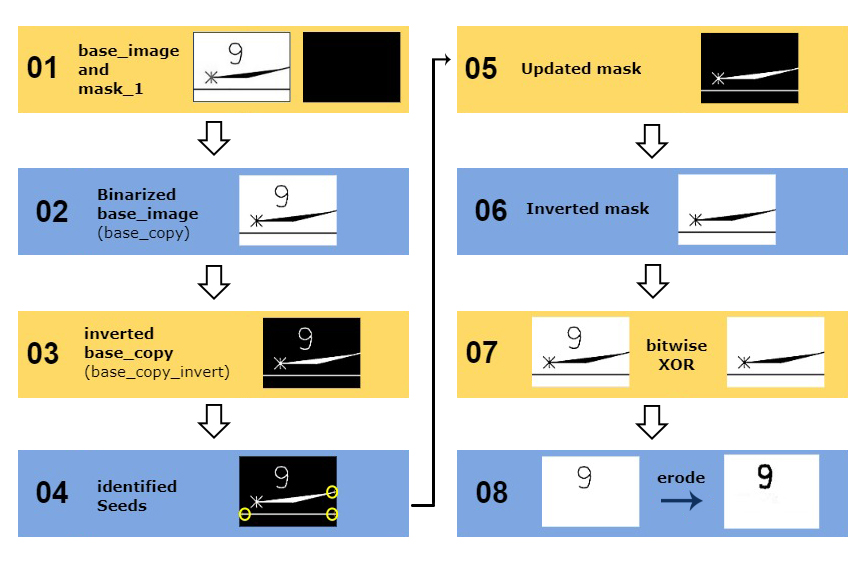}}
  \caption{Distortion Removal Pipeline to enhance OCR (Example)}
  \label{Enhancing-OCR}
\vspace{-0.3cm}
\end{figure}

The need of post processing tasks to enhance the OCR results arises as they also may include additional unnecessary component symbols with railway track portions inside the detected bounding box. To overcome that, below distortion removal pipeline is been used (Figure \ref{Enhancing-OCR}).

\begin{enumerate}
    \item Create a mask of the same size as \textit{base\_image} (\textit{mask\_1})
    \item Make a copy of \textit{base\_image} and convert it into a binary image. (\textit{base\_copy})
    \item Invert the \textit{base\_copy}. Then the components are represented in white pixels [value 1] and the background in black pixels [value 0] (\textit{base\_copy\_inverted})
    \item Provide white pixels that resides at the four edges of the \textit{base\_copy\_inverted} as the seeds for the region growing process. (seeds)
    \item Perform Seeded Region Growing with seeds and base\_copy\_inverted and then store corresponding pixels inside mask\_1. (updated\_mask) inputs : \textit{base\_copy\_inverted}, \textit{mask\_1} and the seeds 
    \item Invert the mask\_1 (\textit{inverted\_mask\_1})
    \item Perform bit-wise XOR operation with \textit{inverted\_mask\_1} and \textit{base\_copy} (\textit{final\_image})
    \item Use morphological enhancements to easily perceive characters (erode white pixels). Then use finalized image as the input image for the OCR process.

\end{enumerate}


\begin{figure}[h]
\vspace{-0.1cm}
  \centering
    \setlength{\fboxsep}{0pt} 
    \setlength{\fboxrule}{0.5pt} 
  \fbox{\includegraphics[width=0.9\linewidth]{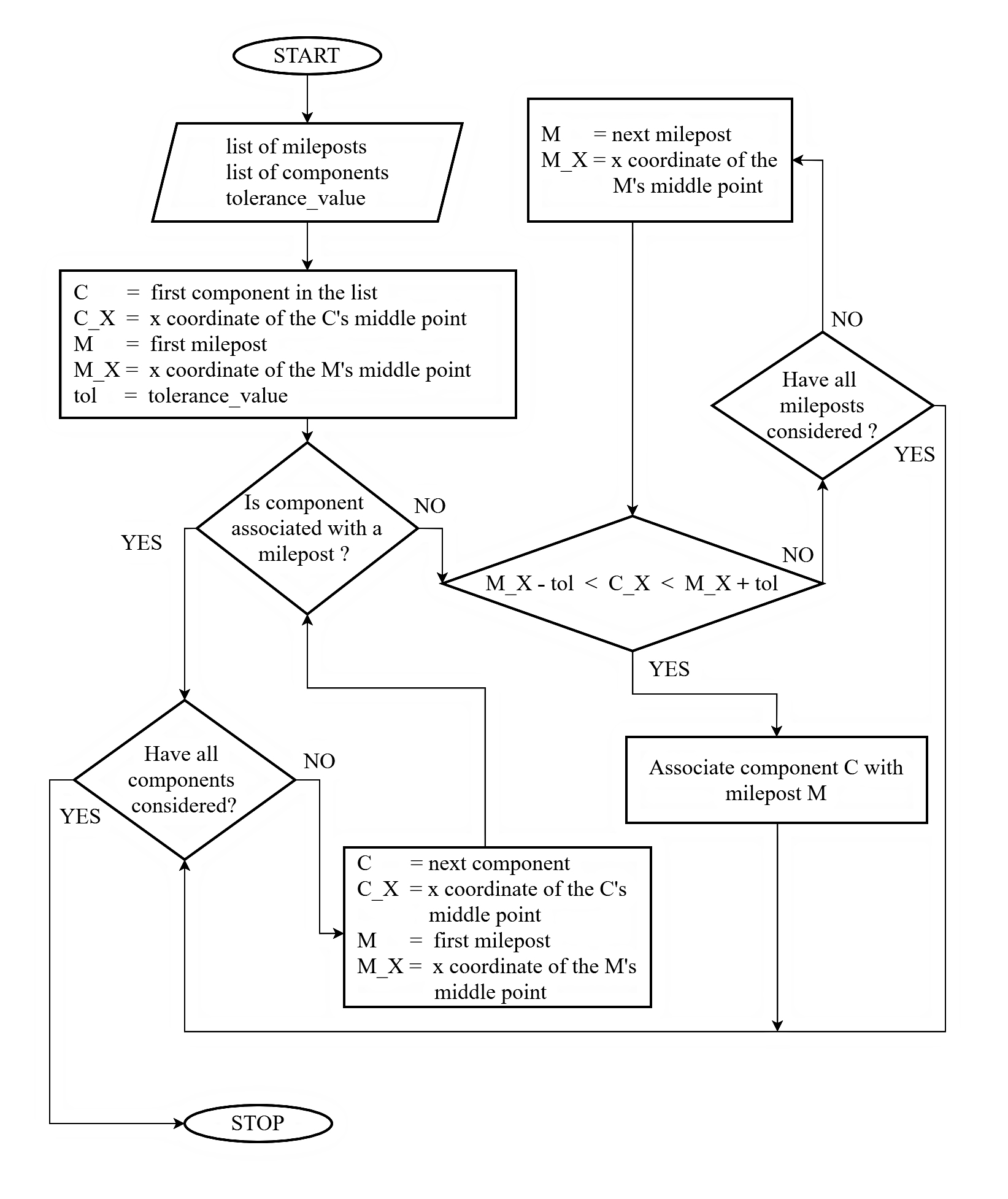}}
  \caption{Solving Component Associativity Problems - Flow Diagram }
  \label{solving_component_mapping_problems}
\vspace{-0.3cm}
\end{figure}

\subsection{Components and text association with corresponding milepost}
Finding which component resides at which milepost is one of the main objectives of this work. Associating detected components with corresponding mileposts was a crucial task, as there are three major problems that needed to be addressed when it came to the process of mapping components with corresponding mileposts (Appendix : Figure \ref{Appendix figure 1}).
\begin{enumerate}
    \item The horizontal area (length) covered by each milepost is not identical to the horizontal area covered (length) by the corresponding component in the input image.
    \item Some milepost areas may have more than one milepost number.
    \item More than one component belongs to the same milepost. That is, a particular clearance point and a signal may have the same milepost number.
\end{enumerate}

The diagram shown in Figure \ref{solving_component_mapping_problems} illustrates the proposed methodology for resolving the problems related with component associativity of mileposts. To overcome the first issue a tolerance value (number of tolerable pixels) is introduced. For the second issue, all milepost values in the considered milepost, have been assigned with each component that satisfies the condition of the solutions for the above problem. To solve the third issue, each component class type is compared with all mileposts.

\subsection{Automated CSV generation process}

A data frame is created to store values for the columns of the form below, and the extracted data are stored and exported as a CSV file accordingly. One CSV file per image is generated automatically without having to explicitly input the images one at a time. Thus the user gets the ability to use the proposed system as a fully automated process. It would only require a directory path containing one or more RTM images as a set of inputs.

\begin{figure}[h]
\vspace{-0.1cm}
  \centering
    \setlength{\fboxsep}{0pt} 
    \setlength{\fboxrule}{0.5pt} 
  \fbox{\includegraphics[width=0.8\linewidth]{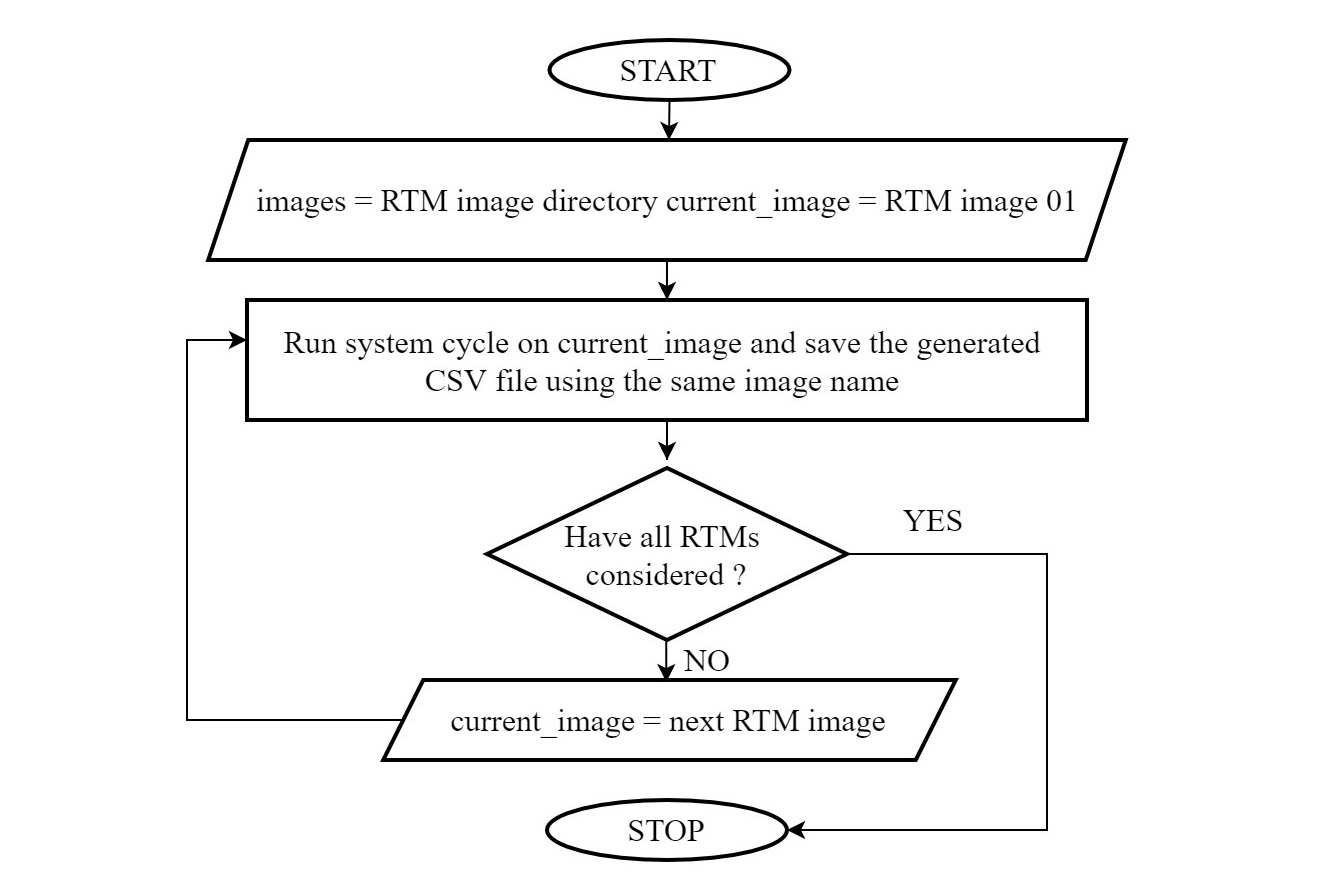}}
  \caption{Solving Component Associativity Problems - Flow Diagram }
  \label{Railway Technical Map}
\vspace{-0.3cm}
\end{figure}

\section{Results And Discussion}

\subsection{Object Detection}

\textbf{Total-Loss for Training phase: }In here it was only possible to achieve a total loss score of 5 as shown in Table 1, which was the best minimum from Faster-RCNN model after 300 epochs of training with 2400 steps having batch size 8.

\textbf{Performance Measures:} Table 2 shows the highest performance measurements per class that were obtained using the Ren Jie Tan's method \cite{towardsdatascienceBreakingDown-ref-5} for Faster-RCNN finalized model. Almost all the F1\_score for each class is above 80\% except for elect-switch class. As the results indicated clearly in Table 3, the exported model from the training job conducted using Faster-RCNN had the overall highest performance values in terms of all mAP, mAR and mAF1 values. 

\begin{table}[ht]
    \caption{Comparison of different losses across models}
    \centering
    \newcolumntype{L}[1]{>{\raggedright\arraybackslash}p{#1}} 
    \newcolumntype{C}[1]{>{\centering\arraybackslash}p{#1}}   
    \newcolumntype{R}[1]{>{\raggedleft\arraybackslash}p{#1}}  
    
    \begin{tabular}{L{5cm}R{1.3cm}R{1.3cm}C{2.4cm}}
    \toprule
    \multicolumn{1}{c}{\textbf{Model}} & \textbf{YOLOv3} & \textbf{SSD} & \textbf{Faster-RCNN}\\
    \textbf{Loss} &  &  &  \\
    \midrule
    \textbf{Confidence loss (conf\_loss)} & 2.34 & 19.41 & 4.31 \\
    
    \textbf{Regression loss (giou\_loss)} & 5.19 & 2.01 & 0.14 \\
    
    \textbf{Classification loss (prob\_loss)} & 2.31 & 7.22 & 1.12 \\
    \toprule
    \textbf{Total loss} & 9.84 & 28.64 & 5.57 \\
    \bottomrule
    \end{tabular}
    \label{table:loss_comparison}
\end{table}

\begin{table}[h!]
    \caption{Final mAP, mAR and mAF1 @IoU\_0.5}
    \centering
    \newcolumntype{L}[1]{>{\raggedright\arraybackslash}p{#1}} 
    \begin{tabular}{L{3cm}ccc}
    \toprule
    \textbf{Model} & \textbf{mAP@IoU\_0.5} & \textbf{mAR@IoU\_0.5} & \textbf{mAF1@IoU\_0.5} \\
    \midrule
    \textbf{YOLOv3} & 0.5013 & 0.6726 & 0.5577 \\
    \textbf{SSD} & 0.4709 & 0.7520 & 0.5646 \\
    \textbf{Faster-RCNN} & 0.6878 & 0.8573 & 0.7618 \\
    \bottomrule
    \end{tabular}
    \label{table:map_mar_maf1}
\end{table}

\begin{table}[h!]
    \caption{Performance measures per-class (Faster-RCNN model) @IoU\_0.5}
    \centering
    
    \newcolumntype{L}[1]{>{\raggedright\arraybackslash}p{#1}} 
    \newcolumntype{C}[1]{>{\centering\arraybackslash}p{#1}}   
    \newcolumntype{R}[1]{>{\raggedleft\arraybackslash}p{#1}}  
    
    \begin{tabular}{L{3cm}R{1cm}R{1cm}R{1cm}R{1cm}R{1cm}R{1cm}}
    \toprule
    \textbf{@IoU\_0.5} & \textbf{TP} & \textbf{FP} & \textbf{FN} & \textbf{AP} & \textbf{AR} & \textbf{F1} \\
    \midrule
    \textbf{milepost} & 77 & 11 & 0 & 0.88 & 1 & 0.93 \\
    \textbf{crossing} & 6 & 3 & 0 & 0.67 & 1 & 0.8 \\
    \textbf{crossing\_label} & 6 & 2 & 0 & 0.75 & 1 & 0.86 \\
    \textbf{signal} & 45 & 14 & 6 & 0.76 & 0.88 & 0.82 \\
    \textbf{switch} & 42 & 12 & 1 & 0.78 & 0.98 & 0.87 \\
    \textbf{clearance\_point} & 48 & 11 & 0 & 0.81 & 1 & 0.9 \\
    \textbf{cp\_name} & 6 & 1 & 0 & 0.86 & 1 & 0.9 \\
    \textbf{elect-switch} & 0 & 1 & 1 & 0 & 0 & 0 \\
    \bottomrule
    \end{tabular}
    \label{table:per_class_performance}
\end{table}

\begin{figure}[htbp]
\vspace{-0.1cm}
  \centering
    \setlength{\fboxsep}{0pt} 
    \setlength{\fboxrule}{0.5pt} 
  \fbox{\includegraphics[width=1\linewidth]{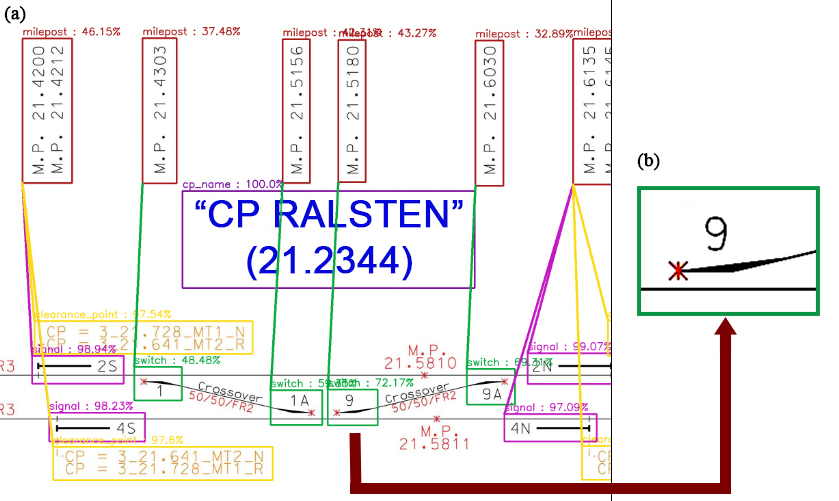}}
  \caption{(a) Identified components associated with corresponding milepost (b) A correctly      labelled switch with unnecessary components for OCR}
  \label{Railway Technical Map}
\vspace{-0.3cm}
\end{figure}

\begin{figure}[htbp]
\vspace{-0.1cm}
  \centering
    \setlength{\fboxsep}{0pt} 
    \setlength{\fboxrule}{0.5pt} 
  \fbox{\includegraphics[width=1\linewidth]{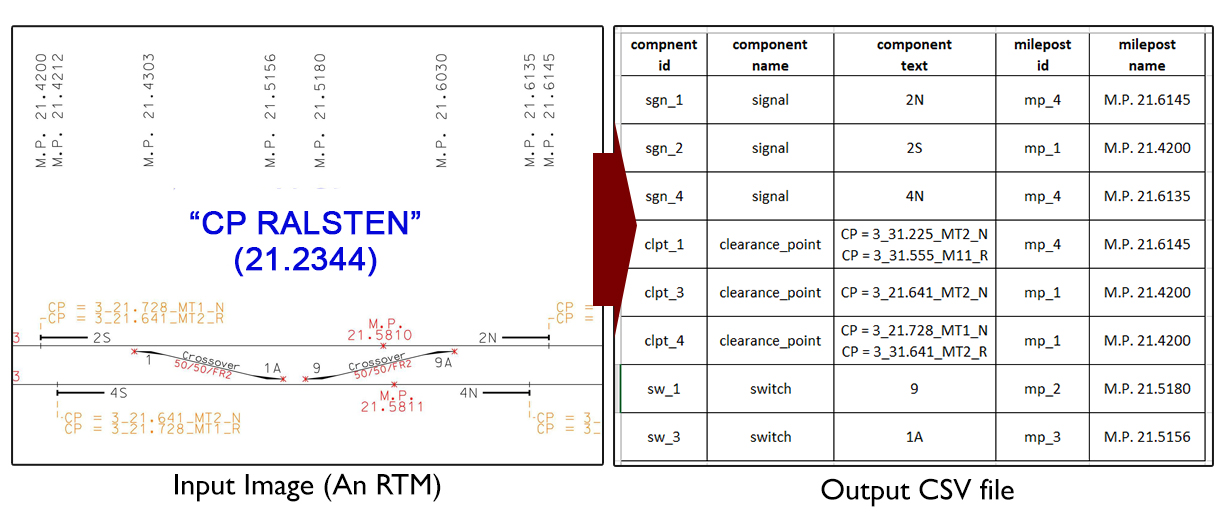}}
  \caption{Final results after character filtering applied}
  \label{Railway Technical Map}
\vspace{-0.3cm}
\end{figure}

\section{Conclusions}
Due to extreme demand for railway transportation, it is important to work on em-bedding digitized mechanisms to make it more effective and efficient. Overall the work carried out in the proposed system, defines and implements an efficient solution that can read RTMs as inputs and produce a formatted string containing all the in-formation extracted from that particular RTM with a restricted training image set (Total of 57: initially 24, with another 33 added later).
The proposed system fully digitizes the extraction of all vital text data related to each RTM component and associating them with corresponding mileposts while resolving the mileposts associativity issues. Further the findings point to the fact that to the current date, Faster-RCNN object detection models are way ahead of YOLOv3 and SSD object detection models in terms of detection accuracy for such applications. 
It is also proven from the results obtained that one can go for a considerably higher accuracy with OCR when the text containing image is sent through a sophisticated problem specific pre-processing pipeline to remove noise and other distortions. When it comes to the future work and insights, the carried out work is involved with pre-identified 8 object types only. However there are more than 15 object types that require different pre-processing techniques to remove distortions inside components before feeding it to the OCR process. Also, finding an optimal associative algorithm is yet to be addressed in future.




\bibliographystyle{unsrt}  
\bibliography{references}  

\clearpage
\section*{Appendix}
Section 3.7: Issues encountered when associating components with milepost markers
\begin{figure}[h]
\vspace{-0.1cm}
  \centering
    \setlength{\fboxsep}{0pt} 
    \setlength{\fboxrule}{0.5pt} 
  \fbox{\includegraphics[width=0.6\linewidth]{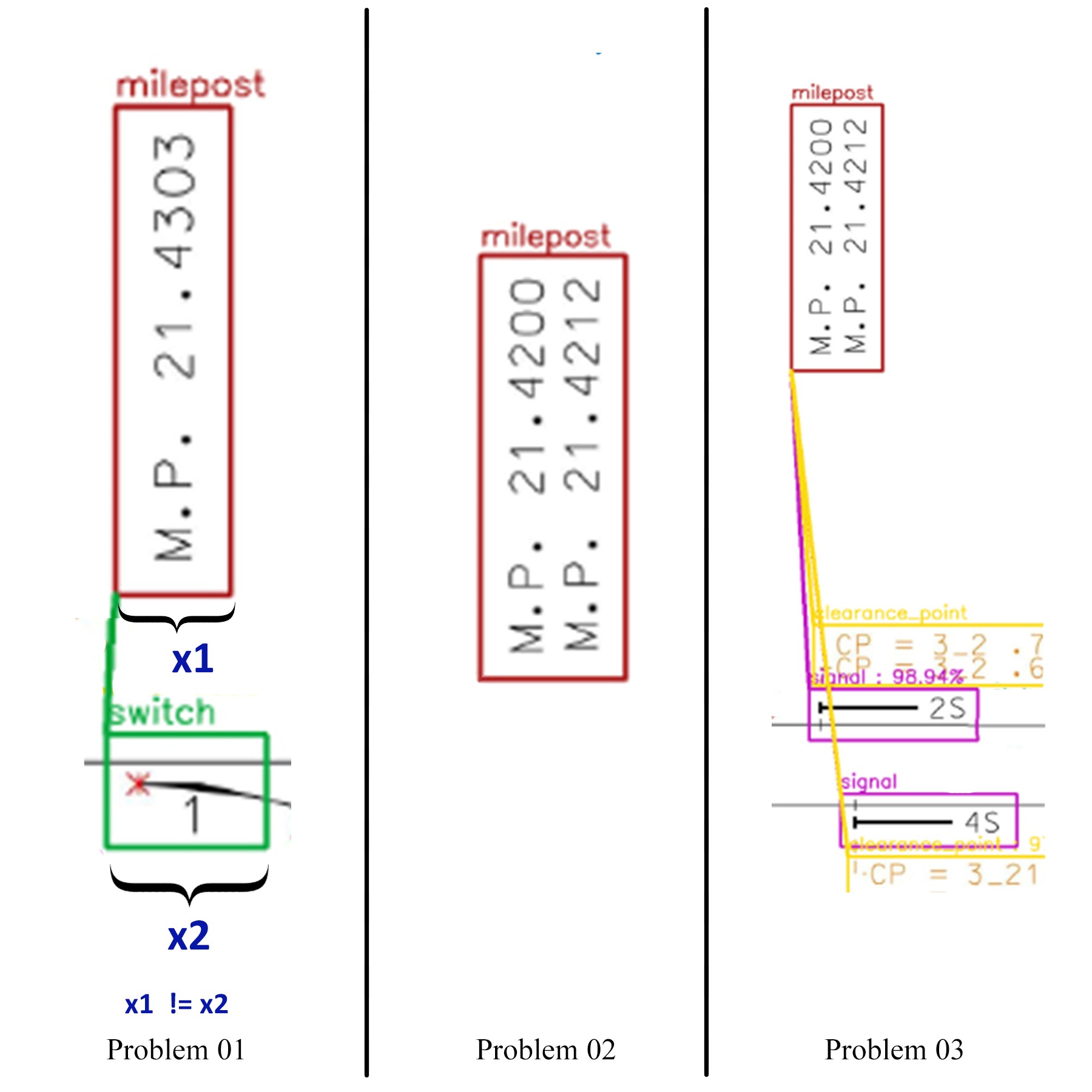}}
  \caption{Issues encountered when associating components with milepost markers}
  \label{Appendix figure 1}
\vspace{-0.3cm}
\end{figure}

\end{document}